\documentclass[12pt]{article}
\usepackage{amsmath}
\usepackage{amssymb}
\usepackage{booktabs}
\usepackage{times}
\usepackage{graphicx}
\usepackage{color}
\usepackage{multirow}
\usepackage{url}
\usepackage[authoryear]{natbib}
\usepackage{rotating}
\usepackage{bbm}
\usepackage{latexsym}

\newcommand{\captionfonts}{\normalsize}

\makeatletter
\long\def\@makecaption#1#2{%
\vskip\abovecaptionskip
\sbox\@tempboxa{{\captionfonts #1: #2}}%
\ifdim \wd\@tempboxa >\hsize
{\captionfonts #1: #2\par}
\else
\hbox to\hsize{\hfil\box\@tempboxa\hfil}%
\fi
\vskip\belowcaptionskip}
\makeatother

\newcommand{\nameOfPaper}{Associated Learning}

\begin{document}
\hspace{13.9cm}1

\ \vspace{20mm}\\

{\LARGE \nameOfPaper{}: Decomposing End-to-end Backpropagation Based on Autoencoders and Target Propagation\footnote{If you are looking for the preprint of the paper published in MIT Neural Computation 33(1) 2021, please see the version 3 on arXiv (\url{https://arxiv.org/abs/1906.05560v3}).  The version you are reading currently includes few more references.}}

\ \\
{\bf \large Yu-Wei Kao, Hung-Hsuan Chen}\\
{Department of Computer Science and Information Engineering, National Central University}\\

{\bf Keywords:} Backpropagation, pipelined training, parallel training, backward locking, associated learning

\thispagestyle{empty}
\markboth{}{NC instructions}
\ \vspace{-0mm}\\
\begin{center} {\bf Abstract} \end{center}

Backpropagation (BP) is the cornerstone of today's deep learning algorithms, but it is
inefficient partially because of backward locking, which means updating the weights of one
layer \emph{locks} the weight updates in the other layers. Consequently, it is
challenging to apply parallel computing or a pipeline structure to update the weights in
different layers simultaneously. In this paper, we introduce a novel learning
structure called associated learning (AL), which modularizes the network into smaller
components, each of which has a local objective. Because the objectives are
mutually independent, AL can learn the parameters in different
layers independently and simultaneously, so it is feasible to apply a 
pipeline structure to improve the training throughput. Specifically, this
pipeline structure improves the complexity of the training time from $O(n
\ell)$, which is the time complexity when using BP and stochastic gradient descent (SGD) for training, to
$O(n+\ell)$, where $n$ is the number of training instances and $\ell$ is the
number of hidden layers. Surprisingly, even though most of the parameters in 
AL do not directly interact with the target variable, training deep
models by this method yields accuracies comparable to those from models trained using
typical BP methods, in which all parameters are used to
predict the target variable. Consequently, because of the scalability 
and the predictive power demonstrated in the experiments, AL
deserves further study to determine the better hyperparameter settings, such as 
activation function selection, learning rate scheduling, and weight initialization,
to accumulate experience, as we have done over the years with the typical BP method. Additionally, perhaps our design can also inspire new network designs 
for deep learning.
Our implementation is available at \url{https://github.com/SamYWK/Associated_Learning}.

\section{Introduction}

Deep neural networks are usually trained using
backpropagation (BP)~\citep{rumelhart1986learning}, which, although common,
increases the training difficulty for several reasons,
among which \emph{backward locking} highly limits the training speed.
Essentially, the end-to-end training method propagates the error-correcting signals layer by
layer; consequently, it cannot update the network parameters of the different layers in
parallel. This backward locking problem is discussed
in~\citep{jaderberg2016decoupled}. Backward locking becomes a severe
performance bottleneck when the network has many layers. Beyond
these computational weaknesses, BP-based learning seems
biologically implausible. For example, it is unlikely that all the weights
would be adjusted sequentially and in small increments based on a single
objective~\citep{crick1989recent}. Additionally, some components 
essential for BP to work correctly have not been observed in the
cortex~\citep{balduzzi2015kickback}. Therefore, many works have proposed
methods that more closely resemble the operations of biological
neurons~\citep{lillicrap2016random, nokland2016direct, bartunov2018assessing,
nokland2019training}. However, empirical studies show that the predictions of
these methods are still unsatisfactory compared to those using
BP~\citep{bartunov2018assessing}.

In this paper, we propose associated learning (AL), a method that can be used to replace
end-to-end BP when training a deep neural network. AL
decomposes the network into small components such that each component has a
local objective function independent of the local objective functions
of the other components. Consequently, the parameters in different components can
be updated simultaneously, meaning that we can leverage parallel computing
or 
pipelining
to improve the training throughput. We conducted experiments on
different datasets to show that AL gives test accuracies
comparable to those obtained by end-to-end BP training, even though most
components in AL do not directly receive the residual signal from the output
layer.

The remainder of this paper is organized as follows. In
Section~\ref{sec:rel-work}, we review the related works regarding the
computational issues of training deep neural networks.
Section~\ref{sec:toy-exp} gives a toy example to compare end-to-end
BP with our proposed AL method. Section~\ref{sec:method}
explains the details of AL. We conducted extensive experiments to
compare AL and BP-based end-to-end learning using
different types of neural networks and different datasets, and the results are
shown in Section~\ref{sec:exp}. Finally, we discuss the discoveries and
suggest future work in Section~\ref{sec:disc}.

\section{Related Work} \label{sec:rel-work}

BP~\citep{rumelhart1986learning} is an essential algorithm for
training deep neural networks and is the foundation of the success of many
models in recent decades~\citep{hochreiter1997long, lecun1998gradient,
he2016deep}. However, because of ``backward locking'' (i.e., the weights must
be updated layer by layer), training a deep neural network can be extremely
inefficient~\citep{jaderberg2016decoupled}. Additionally, empirical evidence
shows that BP is biologically implausible~\citep{crick1989recent,
balduzzi2015kickback, bengio2015towards}. Thus, many studies have suggested
replacing BP with a more biologically plausible method or with a
gradient-free method~\citep{taylor2016training, ororbia2019biologically, ororbia2018conducting} in the hope of decreasing the
computational time and memory consumption and better resembling
biological neural networks~\citep{bengio2015towards, huo2018training,
huo2018decoupled}.

To address the backward locking problem, the authors of~\citep{jaderberg2016decoupled}
proposed using a synthetic gradient, which is an estimation of the real gradient
generated by a separate neural network for each layer. By adopting the
synthetic gradient as the actual gradient, the parameters of every layer can be
updated simultaneously and independently. This approach eliminates the
backward locking problem. However, the experimental results have shown that
this approach tends to result in underfitting---probably because the gradients are difficult
to predict.

It is also possible to eliminate backward locking by computing the local errors
for the different components of a network. In~\citep{belilovsky2018greedy},
the authors showed that using an auxiliary classifier for each layer can yield
good results. However, this paper added one layer to the network at a time, so
it was challenging for the network to learn the parameters of different layers in parallel.
In~\citep{mostafa2018deep}, every layer in a deep neural network is trained by
a local classifier. However, experimental results have shown that this type of model
is not comparable with BP. The authors
of~\citep{belilovsky2019decoupled} and the authors
of~\citep{nokland2019training} also proposed to update parameters based on (or
partially based on) local errors. These models indeed allow the simultaneous updating of
parameters of different layers, and experimental results
showed that these techniques improved testing accuracy. However, these designs
require each local component to receive signals directly from the target
variable for loss computation. Biologically, it is unlikely that neurons
far away from the target would be able to access the target signal directly.
Therefore, even though these methods do not require global BP,
they may still be biologically implausible.

Feedback alignment~\citep{lillicrap2016random} suggests propagating error
signals in a similar manner as BP, but the error signals are propagated with
fixed random weights in every layer. Later, the authors
of~\citep{nokland2016direct} suggested delivering error signals directly from
the output layer using fixed weights. The result is that the gradients are
propagated by weights, while the signals remain local to each layer. The
problem with this approach is that it is similar to the issue discussed in the preceding
paragraph---biologically, distant neurons are unlikely to be able to obtain
signals directly from the target variable.

Another biologically motivated algorithm is target
propagation~\citep{bengio2014auto,
lee2015difference,bartunov2018assessing}. Rather than computing the gradient
for every layer, the target propagation computes the target that each layer
should learn. This approach relies on an
autoencoder~\citep{baldi2012autoencoders} to calculate the inverse mapping of
the forward pass and then pass the ground truth information to every layer.
Each training step includes two losses that must be minimized for each layer:
the loss of inverse mapping and the loss between activations and targets. This
learning method alleviates the need for symmetric weights and is both
biologically plausible and more robust than BP when applied to
stochastic networks. Nonetheless, the targets are still generated layer by
layer.

Overviews of the biologically plausible (or at least partially plausible)
methods are presented in~\citep{bengio2015towards, bartunov2018assessing}.
Although most of these methods perform worse than conventional
BP, optimization beyond BP is still an important
research area, mainly for computational efficiency and biological
compatibility reasons.

Most studies on parallelizing deep learning distribute different data instances into
different computing units. Each of these computing units computes the gradient
based on the allocated instances, and the final gradient is determined by an
aggregation of the gradients computed by all the computing
units~\citep{shallue2018measuring, zinkevich2010parallelized}. Although this
indeed increases the training throughput via parallelization, this is different from
our approach because our method parallelizes the computation in different
layers of a deep network. Our AL technique and the technique of
parallelizing data instances can complement each other and further improve the
throughput given enough computational resources. A recent work, GPipe,
utilizes pipeline training to improve the training throughput~\citep{huang2019gpipe}.
However, all the parameters in GPipe are still influenced in a layerwise
fashion. Our method is different because the parameters in the different layers
are independent.

Our work is highly motivated by target propagation, but we create
intermediate mappings instead of directly transforming features into targets.
As a result, the local signals in each layer are independent of the signals
in the other layers, and most of these signals are not obtained directly from
the output label.

\section{A Toy Example to Compare the Training Throughput of End-to-end Backpropagation and \nameOfPaper{}}
\label{sec:toy-exp}

\begin{figure}[tb]
\begin{center}
\centerline{\includegraphics[angle=270,width=0.8\textwidth]{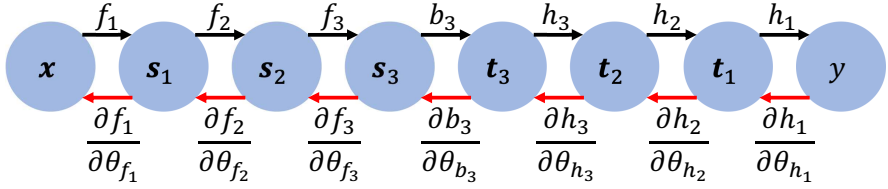}}
\caption{
An example of a deep neural network with 6 hidden layers.
We denote each forward function ($f_1, f_2, f_3, b_3, h_3, h_2, h_1$) and
the output of each function ($s_1, s_2, s_3, t_3, t_2, t_1, y$) by different symbols
for ease of later explanation. Let $\theta^{(f)}$ denote the parameters of 
a function $f$; then, the backward path requires computing the local gradient
$\frac{\partial f}{\partial \theta^{(f)}}$ for each function $f$.
}
\label{fig:dnn-structure}
\end{center}
\end{figure}

Figure~\ref{fig:dnn-structure} gives a typical structure of a deep neural
network with 6 hidden layers. The input feature vector $\boldsymbol{x}$ goes
through a series of transformations ($\boldsymbol{x} \xrightarrow{f_1}
\boldsymbol{s}_1 \xrightarrow{f_2} \boldsymbol{s}_2 \xrightarrow{f_3}
\boldsymbol{s}_3 \xrightarrow{b_3} \boldsymbol{t}_3 \xrightarrow{h_3}
\boldsymbol{t}_2 \xrightarrow{h_2} \boldsymbol{t}_1 \xrightarrow{h_1} y$) to
approximate the corresponding output $y$. We denote the functions ($f_1, f_2,
f_3, b_3, t_3, t_2, t_1$) and the outputs of these functions ($s_1, s_2, s_3, t_3,
t_2, t_1, y$) by different symbols for the ease of later explanation on
AL. If stochastic gradient descent (SGD) and BP are applied to
search for the proper parameter values, we need to compute the local gradient
$\frac{\partial f}{\partial \theta^{(f)}}$ as the backward function for every
forward function $f$ (whose parameters are denoted by $\theta^{(f)}$). As a
result, each training epoch requires a time complexity of $O(n \times ((\ell+1) +
(\ell+1))) \approx O(n \ell)$, in which $n$ is the number of training
instances and $\ell$ is the number of hidden layers (i.e., $\ell=6$ in our
example). Since both forward pass and backward pass require $\ell +
1$ transformations, we have two $\ell + 1$ terms. Consequently, the training
time increases linearly with the number of hidden layers $\ell$.

\begin{figure}[tb]
\begin{center}
\centerline{\includegraphics[angle=270,width=0.6\linewidth]{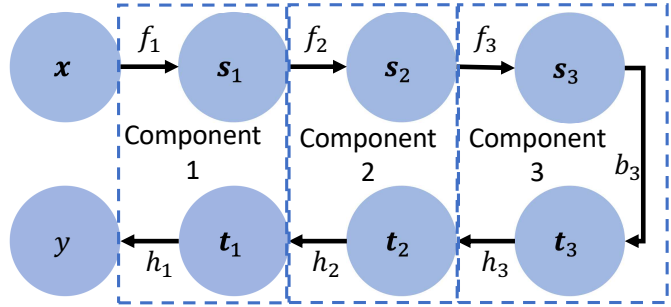}}
\caption{
A simplified structure of the AL technique, which decomposes 6 hidden layers
into 3 components such that each component has a local objective function that
is independent of the objective functions of the other components. Consequently, we
may update the parameters in component $i$ $(\theta^{(f)}_i, \theta^{(h)}_i)$ and the parameters
in component $j$ $(\theta^{(f)}_j, \theta^{(h)}_j)$ simultaneously for $i \neq j$.
}
\label{fig:al-simple-structure}
\end{center}
\end{figure}

\begin{table}[tbh]
\caption{An example of simultaneously updating the parameters by pipelining}
\label{tab:pipeline}
\resizebox{\columnwidth}{!}{
\begin{tabular}{l||llllllll}
\toprule
Time unit      & 1      & 2      & 3      & 4      & 5      & 6      & 7      & ...\\
\midrule
$1^{\textrm{st}}$ mini-batch & Task 1 & Task 2 & Task 3 &        &        &        &        & \\
$2^{\textrm{nd}}$ mini-batch &        & Task 1 & Task 2 & Task 3 &        &        &        & \\
$3^{\textrm{rd}}$ mini-batch &        &        & Task 1 & Task 2 & Task 3 &        &        & \\
$4^{\textrm{th}}$ mini-batch &        &        &        & Task 1 & Task 2 & Task 3 &        & \\
$5^{\textrm{th}}$ mini-batch &        &        &        &        & Task 1 & Task 2 & Task 3 & \\
...            &        &        &        &        &        &        &        & \\
\bottomrule
\end{tabular}
}
\end{table}

Figure~\ref{fig:al-simple-structure} shows a simplified structure of
the AL technique, which ``folds'' the network and decomposes the network into 3 components such that each component has a local objective function that is
independent of the local objectives in the other components. As a result, for $i
\neq j$, we may update the parameters in component $i$ $(\theta^{(f)}_i, \theta^{(h)}_i)$ and the parameters in component $j$ $(\theta^{(f)}_j, \theta^{(h)}_j)$ independently and simultaneously, since the parameters of component $i$ ($(\theta^{(f)}_i, \theta^{(h)}_i)$) determine the loss of component $i$, which is independent of the loss of component $j$, which is
determined by the parameters of component $j$ ($(\theta^{(f)}_j, \theta^{(h)}_j)$). 

Table~\ref{tab:pipeline} gives an example of applying pipelining for parameter updating to improve the training throughput using AL. Let Task
$i$ be the task of updating the parameters in Component $i$. At the $1^{\textrm{st}}$ time
unit, the network performs Task 1 (updating $\theta^{(f)}_1$ and $\theta^{(h)}_1$)
based on the $1^{\textrm{st}}$ training instance (or the instances in the $1^{\textrm{st}}$ mini-batch). At
the $2^{\textrm{nd}}$ time unit, the network performs Task 1 (updating $\theta^{(f)}_1$ and
$\theta^{(h)}_1$) based on the $2^{\textrm{nd}}$ training instance (or the training instances in
the 2nd mini-batch) and performs Task 2 (updating $\theta^{(f)}_2$ and
$\theta^{(h)}_2$) based on the $1^{\textrm{st}}$ instance (or the $1^{\textrm{st}}$ mini-batch). As shown in the table,
starting from the $3^{\textrm{rd}}$ time unit, the parameters in all the different components
can be updated simultaneously. Consequently,
the first instance requires $O(\ell/2)$ units of computational time, and, because
of the pipeline, each of the following $n-1$ instances requires only $O(1)$ 
units of computational time. Therefore, the time complexity of each training epoch
becomes $O(\ell/2 + (n-1)) \approx O(n+\ell)$.

Compared to end-to-end BP during which the time complexity grows
linearly to the number of hidden layers, the time complexity of the proposed
AL with pipelining technique grows to only a constant time as the number of hidden layers increases.

\section{Methodology} \label{sec:method}

\begin{figure}[tb]
\begin{center}
\centerline{\includegraphics[angle=270,width=0.6\linewidth]{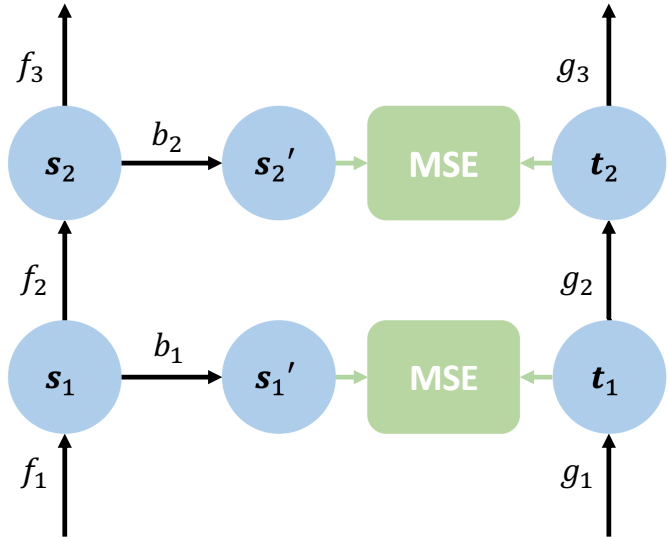}}
\caption{Adding a ``bridge'' to the structure. The bridge includes nonlinear
layers to transform $s_i$ into $s_i'$ such that $s_i' \approx t_i$.
The black arrows indicate the forward path.}
\label{fig:substructure}
\end{center}
\end{figure}

A typical deep network training process requires features to pass through
multiple nonlinear layers, allowing the output to approach the ground-truth
labels. Therefore, there is only one objective. With AL, however, we
modularize the training path by splitting it into smaller components and assign
independent local objectives to each small component. Consequently, the
AL technique divides the original long gradient flow into many independent
short gradient flows and effectively eliminates the backward locking problem.
In this section, we introduce three types of functions (associated function,
encoding and decoding functions, and bridge function) that together compose
the AL network.

\subsection{Associated Function and Associated Loss}

Referring to Figure~\ref{fig:al-simple-structure}, let $\boldsymbol{x}$ and $y$
be the input features and the output target, respectively, of a training sample. We split a
network with $\ell$ hidden layers into $\ell/2$ components (assuming $\ell$ is
an even number). The details of each component are illustrated in Figure~\ref{fig:asso-learn-all}. Each component $i$ consists of two local forward
functions, $f_i$ and $g_i$ ($f_i$ and $g_i$ will be called the \emph{associated
function} and \emph{encoding function}, respectively, for better differentiation; we will further 
explain the encoding function in Section~\ref{sec:encode-decode-func}), and
a local objective function independent of the objective functions of
the other components. A local associated function can be a simple single-layer
perceptron, a convolutional layer, or another function. We compute $s_i$ using
Equation~\ref{eq:x-forward}:

\begin{equation} \label{eq:x-forward}
s_i = f_i(s_{i-1}),\textrm{ } i = 1, \ldots, \ell/2.
\end{equation}
Note that here, $s_0$ equals $\boldsymbol{x}$.

We define the \textit{associated loss function} for each pair of $(s_i, t_i)$
by Equation~\ref{eq:forward-loss}. This concept is similar
to target propagation~\citep{bengio2014auto, lee2015difference,
bartunov2018assessing}, in which the goal is to minimize the distance between $s_i$
and $t_i$ for every component $i$.

\begin{equation} \label{eq:forward-loss}
L_i(s_i, t_i)=||s_i-t_i||^2,\textrm{ } i = 1, \ldots, \ell/2.
\end{equation}

The optimizer in the $i^{\textrm{th}}$ component updates the parameters in
$f_i$ to reduce the associated loss function (Equation~\ref{eq:forward-loss}).

Referring to Figure~\ref{fig:al-simple-structure},
Equation~\ref{eq:forward-loss} attempts to make $\boldsymbol{s}_i \approx
\boldsymbol{t}_i$ for all $i$. This design may look strange for several reasons.
First, if we can obtain an $f_1$ such that $\boldsymbol{s}_1 \approx
\boldsymbol{t}_1$, all the other $f_i$s $(i > 1)$ seem unnecessary.
Second, since $\boldsymbol{s}_1$ and $\boldsymbol{t}_1$ are far apart,
fitting these two terms seems counterintuitive.

For the first question, one can regard each component as one layer in a deep
neural network. As we add more components, the corresponding $\boldsymbol{s}_i$ and
$\boldsymbol{t}_i$ may become closer. For the second question, indeed, it seems
more reasonable to fit the values of neighboring cells. However, our design
breaks the gradient flow among different components so that it is possible to
perform a parallel parameter update for each component.

\subsection{Bridge Function}  \label{sec:reformedAL}

Our early experiments showed that $\boldsymbol{s}_i$ has difficulty fitting the
corresponding target $\boldsymbol{t}_i$, especially for a convolutional
neural network (CNN) and its variants. Thus, we insert nonlinear layers to improve
the fitting between $\boldsymbol{s}_i$ and $\boldsymbol{t}_i$. As shown in
Figure~\ref{fig:substructure}, we create a \emph{bridge function}, $b_i$, to
perform a nonlinear transform on $\boldsymbol{s}_i$ such that
$b_i(\boldsymbol{s}_i) = \boldsymbol{s}_i' \approx \boldsymbol{t}_i$. As a
result, the associated loss is reformulated to the following equation to
replace the original Equation~\ref{eq:forward-loss}:

\begin{equation} \label{eq:new-forward-loss}
L_i(\boldsymbol{s}_i, \boldsymbol{t}_i)=||b_i(\boldsymbol{s}_i)-\boldsymbol{t}_i||^2,\textrm{ } i = 1, \ldots, \ell/2,
\end{equation}
where the function $b_i(.)$ serves as the bridge.

Although this approach greatly increases the number of parameters and the nonlinear layers to
decrease the forward loss, except for the last bridge, these parameters do not
affect the inference function, as we will explain in
Section~\ref{sec:predictingPhase}, so the bridges only slightly increase the
hypothesis space. For a fair comparison, we also increase the number of
parameters when the models are trained by BP so that the models
trained by AL and trained by BP have the same number 
of parameters. The details will be explained in Section~\ref{sec:exp}.

\subsection{Encoding/Decoding Functions and Autoencoder Loss} \label{sec:encode-decode-func}

Referring to Figure~\ref{fig:al-simple-structure}, in addition to the
parameters of the $f_i$s and $b_i$s, we also need to obtain parameters in $h_i$s to
have the mapping $\boldsymbol{t}_i \rightarrow \boldsymbol{t}_{i-1}$ at the inference phase. This mapping is
achieved by the following two functions, which together can be regarded as an
autoencoder:

\begin{equation} \label{eq:y-encode}
\boldsymbol{t}_i = g_i(\boldsymbol{t}_{i-1}),~i = 1, \ldots, \ell/2.
\end{equation}

\begin{equation} \label{y-decode}
\boldsymbol{t}'_{i-1}=h_i(\boldsymbol{t}_i),~ i = i, \ldots, \ell/2.
\end{equation}

Referring to Figure~\ref{fig:asso-learn-all}, the above two equations form an
autoencoder because we want $\boldsymbol{t}_{i-1} \xrightarrow{g_i} \boldsymbol{t}_i \xrightarrow{h_i}
\boldsymbol{t}_{i-1}' \approx \boldsymbol{t}_{i-1}$, so $g_i$ and $h_i$ are called the  \emph{encoding
function} and \emph{decoding function}, respectively. The autoencoder loss
$L'_i$ for layer $i$ is defined by Equation~\ref{eq:autoencoder-loss}:

\begin{equation} \label{eq:autoencoder-loss}
L_i'(h_i(g_i(\boldsymbol{t}_{i-1})), \boldsymbol{t}_{i-1}) = ||\boldsymbol{t}_{i-1}'-\boldsymbol{t}_{i-1}||^2, i = 1, \ldots ,\ell/2.
\end{equation}

\subsection{Putting Everything Together}

\begin{figure}[tb]
\begin{center}
\centerline{\includegraphics[angle=270,width=\linewidth]{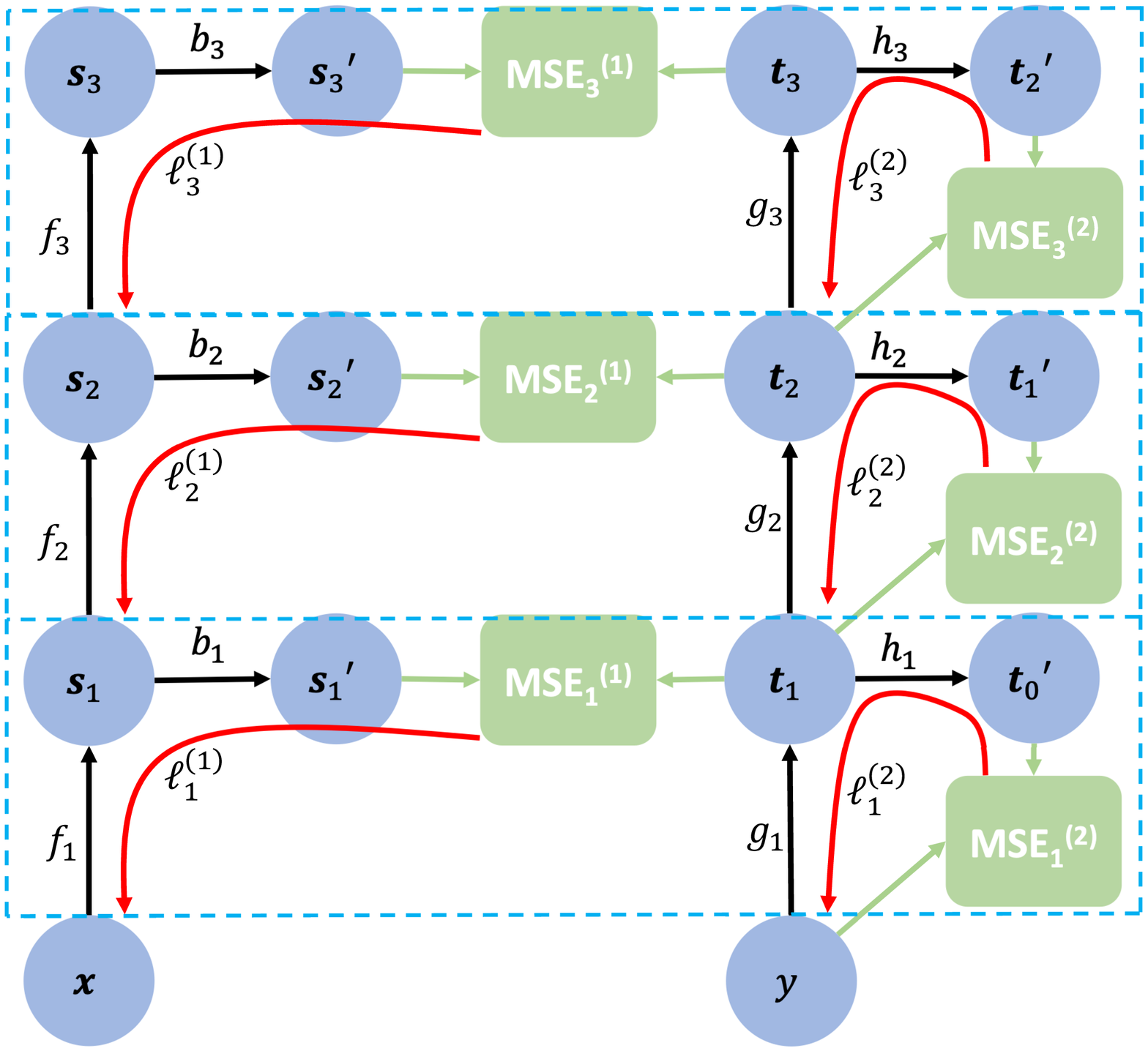}}
\caption{A training example using associated learning. The black arrows indicate
the forward paths that involve learnable parameters; the green arrows connect
the variables that should be compared to minimize their associated distance; the red
arrows denote the backward gradient flows. We group each component by dashed
lines. The parameters of the different components are independent so that they
can be updated simultaneously. The variable $\ell_u^{(v)}$ denotes the $v$th
gradient flow of the $u$th component. $\textrm{MSE}_u^{(v)}$ denotes the
$v$th mean-squared error of the $u$th component. Consequently, the first
gradient flow of each component, $\ell_u^{(1)}$, determines the updates of the
parameters of $f_u$ and $b_u$; the second gradient flow of each component,
$\ell_u^{(2)}$, determines the updates of $g_u$ and $h_u$.
}
\label{fig:asso-learn-all}
\end{center}
\end{figure}

Figure~\ref{fig:asso-learn-all} shows the entire training process of
AL based on our earlier example. We group each component by a 
dashed line. The parameters in each component are independent of the
parameters in the other components. For each component $i$, the local objective
function is defined by Equation~\ref{eq:local-obj}.

\begin{equation} \label{eq:local-obj}
\textrm{local-obj}_i
= \textrm{MSE}_i^{(1)} + \textrm{MSE}_i^{(2)}
= ||b_i(\boldsymbol{s}_i) - \boldsymbol{t}_i||^2 + ||\boldsymbol{t}_{i-1}' - \boldsymbol{t}_{i-1}||^2,
\end{equation}
where $||b_i(\boldsymbol{s}_i) - \boldsymbol{t}_i||^2$ is the associated loss shown by
Equation~\ref{eq:new-forward-loss} and $||\boldsymbol{t}_{i-1}' - \boldsymbol{t}_{i-1}||^2$ is the autoencoder loss demonstrated by Equation~\ref{eq:autoencoder-loss}.

As shown in Figure~\ref{fig:asso-learn-all}, the associated loss in each
component creates the gradient flow $\ell_i^{(1)}$, which guides the updates of
the parameters of $f_i$ and $b_i$. The autoencoder loss in each component
leads to the second gradient flow $\ell_i^{(2)}$, which determines the updates
of $g_i$ and $h_i$.

A gradient flow travels only within a component, so the parameters in
different components can be updated simultaneously. Additionally, since each
gradient flow is short, the vanishing gradient and exploding gradient problems
are less likely to occur.

Since each component incrementally refines the association loss of the component 
immediately below it, 
the input $\boldsymbol{x}$ approaches the output $y$.

\subsection{Inference Function, Effective Parameters, and Hypothesis Space}  \label{sec:predictingPhase}

We can categorize the abovementioned parameters into two types: effective
parameters and affiliated parameters. The affiliated parameters help the
model determine the values of the effective parameters, which in turn determine
the hypothesis space of the final inference function. Therefore, while
increasing the number of affiliated parameters may help to obtain better
values for the effective parameters, it will not increase the hypothesis
space of the prediction model. Such a setting may be relevant to the
overparameterization technique, which introduces redundant parameters to
accelerate the training speed~\citep{allen2018convergence, arora2018optimization,
chen2017weighted, chen20acclerating}, but here, the purpose is to obtain better values of the
effective parameters rather than faster convergence.

Specifically, in the training phase, we search for the parameters of the $f_i$s and
$b_i$s that minimize the associated loss and search for the parameters of the 
$g_i$s and $h_i$s to minimize the autoencoder loss. However, in the inference
phase, we make predictions based only on Equation~\ref{eq:x-forward},
Equation~\ref{y-decode}, and $b_{\ell/2}(\boldsymbol{s}_{\ell/2})$. Therefore, the effective parameters
include only the parameters in the $f_i$s, the $h_i$s ($i=1,\ldots,\ell/2$), and $b_{\ell/2}$
(i.e., the last bridge). The parameters in the other functions (i.e., the
$g_i$s $(i=1, \ldots, \ell/2)$ and the $b_j$s $(j=1,\ldots, \ell/2-1)$) are affiliated parameters; they do not
increase the expressiveness of the model but only help determine the values of
the effective parameters.

The predicting process can be represented in
Figure~\ref{fig:al-simple-structure}. Equation~\ref{eq:pred-func} shows the
prediction function:

\begin{equation} \label{eq:pred-func}
\hat{y} = \left(h_1 \circ h_2 \circ \ldots \circ h_{\ell/2} \circ b_{\ell/2} \circ f_{\ell/2} \circ \ldots \circ f_2 \circ f_1\right)(\boldsymbol{x}),
\end{equation}
where $\circ$ denotes the function composition operation and $\ell=6$ in the
example is illustrated by Figure~\ref{fig:al-simple-structure} and
Figure~\ref{fig:asso-learn-all}. Only the parameters involved in
Equation~\ref{eq:pred-func} are the effective parameters that determine the
hypothesis space.

\section{Experiments}  \label{sec:exp}

In this section, we introduce the experimental settings, implementation details,
and show the results of the performance comparisons between
BP and AL.

\subsection{Experimental Settings}

We conducted experiments by applying AL and BP to different deep neural
network structures (a multilayer perceptron (MLP), a vanilla CNN, a Visual Geometry Group (VGG) network~\citep{simonyan15very},
a 20-layer residual neural network (ResNet-20), and a 32-layer ResNet (ResNet-32)~\citep{he2016deep}) and different datasets (the Modified National Institute of Standards and Technology 
(MNIST)~\citep{lecun1998gradient}, the 10-class Canadian Institute for Advanced Research (CIFAR-10), and the 100-class CIFAR 
(CIFAR-100)~\citep{krizhevsky2009learning} datasets). Surprisingly, although the AL approach aims at minimizing the local losses, its prediction accuracy is comparable to, and
sometimes even better than, that of BP-based learning, whose goal
is directly minimizing the prediction error.

In each experiment, we used the settings that were reported in recent papers.
We spent a reasonable amount of time searching for the hyperparameters not
stated in previous papers based on random search~\citep{bergstra2012random}.
Eventually, we initialized all the weights based on the He normal initializer
and use Adam as the optimizer. We experimented with different activation
functions and adopted the exponential linear unit (ELU) for all the local forward functions (i.e., $f_i$) and a 
sigmoid function for the functions related to the autoencoders and bridges (i.e., $g_i$,
$h_i$, and $b_i$).
The models trained by BP yielded test accuracies close to the state-of-the-art (SOTA) results
under the same or similar network structures~\citep{he2016deep,
carranza2019unsharp}.
In addition, because AL includes extra parameters in the function
$b_{\ell/2}$ (the last bridge), as explained in
Section~\ref{sec:predictingPhase}, we increased the number of layers in the
corresponding baseline models when training by BP so that
the models trained by AL and those trained by BP have
identical parameters, so the comparisons are fair.

The implementations are freely available at \url{https://github.com/SamYWK/Associated_Learning}.

\subsection{Test Accuracy}

\begin{table}
\caption{Test accuracy comparison on the MNIST dataset. We
highlight the winner in bold font. We applied only the 
DTP algorithm on the MLP
because this is the setting used in the original paper. Applying DTP
on other networks might require different designs.
}
\label{tab:mnist}
\centering
\begin{tabular}{l||ccc}
\toprule
            & BP               &  AL  & DTP \\
\midrule
MLP         & $98.5 \pm 0.0\%$ & $\boldsymbol{98.6 \pm 0.0\%}$ & $96.43 \pm 0.04\%$ \\
Vanilla CNN & $99.4 \pm 0.0\%$ & $\boldsymbol{99.5 \pm 0.0\%}$ & - \\
\bottomrule
\end{tabular}
\end{table}

To test the capability of AL, we compared AL and BP on different network 
structures (MLP, vanilla CNN, ResNet, and VGG) and different datasets (MNIST,
CIFAR-10, and CIFAR-100). 
When converting a network with an odd number of layers into the "folded"
architecture used by AL, the middle layer is simply absorbed by the bridge
layer at the top component shown in Figure~\ref{fig:asso-learn-all}.
We also experimented with differential target propagation
(DTP)~\citep{lee2015difference} on the MLP network based on the MNIST dataset.
We tried only the MLP network, as the original paper applied only DTP to the MLP
structure and applying DTP to other network structures requires different designs.

On the MNIST dataset, we conducted experiments with only two networks
structures, MLP and vanilla CNN, because using even these simple structures
yielded decent test accuracies. Their detailed settings
are described in the following paragraphs.
The results are shown in Table~\ref{tab:mnist}. For both the MLP and the
vanilla CNN structure, AL performs slightly better than 
BP, which performs better than DTP on the MLP network.

The MLP contains 5 hidden layers and 1 output layer; there are $1024$, $1024$, $5120$, $1024$, and $1024$ neurons in the hidden layers and $10$ neurons in the output layer. Referring to Figure~\ref{fig:asso-learn-all}, this network
corresponds to the following structure when using the AL framework: the network
has two components; both the $\boldsymbol{s}_i$ and $\boldsymbol{t}_i$ in a
component $i$ ($i=1,2$) have $1024$ neurons, and $\boldsymbol{b}_2$ the output of the top
bridge function contains $5120$ neurons.

The vanilla CNN contains 13 hidden layers and 1 output layer. The first 4 layers  are
convolutional layers with a size of $3 \times 3 \times 32$ (i.e., a width of 3, a 
height of 3, and 32 kernels) in each layer, followed by 4 convolutional layers
with a size of $3 \times 3 \times 64$ in each layer, followed by a
fully connected layer with $1280$ neurons, followed by 4 fully connected layers
with $256$ neurons in each layer and ending with a fully connected layer with
10 neurons. When training by AL, this structure corresponds to the following:
the first five layers (layers 1 to 5) and the last five layers (layers 9 to 13)
form five components, where layer $i$ and layer $14 - i$ ($i=1, \ldots, 5$)
belong to component $i$ and the $6^{\textrm{th}}$, $7^{\textrm{th}}$, and
$8^{\textrm{th}}$ layers construct the component $6$. The initial learning rate is
$10^{-4}$, which is reduced after $80$, $120$, $160$, and $180$ epochs.

\begin{table}
\caption{Test accuracy comparison on the CIFAR-10 dataset. We highlight the winner in bold font. We applied only the DTP algorithm on the MLP because this is the setting used in the original paper. Applying DTP
on other networks might require different designs.
}
\label{tab:cifar10}
\centering
\begin{tabular}{l||ccc}
\toprule
            & BP                            & AL  & DTP\\
\midrule 
MLP         & $60.6 \pm 0.3\%$              & $\boldsymbol{62.8 \pm 0.2\%}$ & $58.2 \pm 0.2\%$ \\
Vanilla CNN & $85.2 \pm 0.4\%$              & $\boldsymbol{85.8 \pm 0.1\%}$ & - \\
ResNet-20   & $\boldsymbol{91.2 \pm 0.4\%}$ & $89.1 \pm 0.5\%$ & - \\
ResNet-32   & $\boldsymbol{92.0 \pm 0.2\%}$ & $88.7 \pm 0.4\%$ & - \\
VGG         & $92.3 \pm 0.2\%$              & $\boldsymbol{92.6 \pm 0.1\%}$ & - \\
\bottomrule
\end{tabular}
\end{table}

\begin{table}
\caption{Test accuracy comparison on the CIFAR-100 dataset. We highlight the winner in bold font.}
\label{tab:cifar100}
\centering
\begin{tabular}{l||cc}
\toprule
            & BP                            & AL  \\
\midrule
MLP         & $26.5 \pm 0.4\%$              & $\boldsymbol{29.7 \pm 0.2\%}$ \\
Vanilla CNN & $51.1 \pm 0.2\%$              & $\boldsymbol{52.2 \pm 0.5\%}$ \\
ResNet-20   & $\boldsymbol{63.7 \pm 0.2\%}$ & $61.0 \pm 0.6\%$ \\
ResNet-32   & $\boldsymbol{63.7 \pm 0.3\%}$ & $59.0 \pm 1.6\%$ \\
VGG         & $65.8 \pm 0.3\%$              & $\boldsymbol{67.1 \pm 0.3\%}$ \\
\bottomrule
\end{tabular}
\end{table}

The CIFAR-10 dataset is more challenging than the MNIST dataset. The input
image size is $32\times32\times3$~\citep{krizhevsky2009learning}; i.e., the
images have a higher resolution, and each pixel includes red, green, and blue (RGB) information. To make good use of these abundant features, we included not only MLP and vanilla
CNN in this experiment but also VGG and the ResNets. The input images are augmented
by 2-pixel jittering~\citep{sabour2017dynamic}. We applied the L2-norm using
$5 \times 10^{-4}$  and $1\times 10^{-4}$  as the regularization
weights for VGG and the ResNet models.

Because ResNet uses batch normalization and the shortcut trick, we set its
learning rate to $10^{-3}$, which slightly larger than that of the other models. In addition, to
ensure that the models trained by BP and AL have identical numbers of parameters
for a fair comparison, we added extra layers to ResNet-20, ResNet-32, and VGG
when using BP for learning.

Table~\ref{tab:cifar10} shows the results of the CIFAR-10 dataset.
AL performs marginally better than BP on the MLP,
vanilla CNN, and VGG structures. With the ResNet structure, AL
performs slightly worse than BP. The CIFAR-100 dataset includes
100 classes. We used model settings that were nearly identical to the settings
used on the CIFAR-10 dataset but increased the number of neurons in the bridge.
Table~\ref{tab:cifar100} shows the results. As in CIFAR-10, AL 
performs better than BP on the MLP, vanilla CNN, and VGG structures but slightly worse
on the ResNet structures.

Currently, the theoretical aspects of the AL method are weak, so we are unsure
of the fundamental reasons why AL outperforms BP on MLP, vanilla
CNN, and VGG but BP outperforms AL on ResNet.  Our speculations are below.
First, since BP aims to fit the target directly, and most of the
layers in AL can leverage only indirect clues to update the
parameters, AL is less likely to outperform BP.
However, this reason does not explain why AL performs better than BP on
other networks. Second, perhaps the bridges can be regarded implicitly as the
shortcut connections of ResNet, so applying AL on ResNet appears
such as refining residuals of residuals, which could be noisy. Finally, years of
study on BP has made us gain experience on the hyperparameter
settings for BP. A similar hyperparameter setting may not
necessarily achieve the best setting for AL.

As reported in~\citep{bartunov2018assessing}, earlier studies on
BP alternatives, such as target propagation (TP) and feedback
alignment (FA), performed worse than BP in
non-fully connected networks (e.g., a locally connected network such as a CNN)
and more complex datasets (e.g., CIFAR). Recent studies, such as those on decoupled greedy learning (DGL) and
the Predsim model~\citep{belilovsky2019decoupled, nokland2019training}, showed
a similar performance to BP on more complex networks, e.g., VGG, but
these models require each layer to access the target label $y$ directly, which
could be biologically implausible because distant neurons are unlikely to
obtain the signals directly from the target. As far as we know, our proposed
AL technique is the first work to show that an alternative of BP
works on various network structures without directly revealing the target $y$
to each hidden layer, and the results are comparable to, and sometimes even
better than, the networks trained by BP.

\subsection{Number of Layers vs. the Associated Loss and vs. the Accuracy}

\begin{table}
\caption{The associated loss at different layers on the MNIST dataset after 200 epochs.
Referring to Figure~\ref{fig:asso-learn-all}, for each layer,
its corresponding $\boldsymbol{s}_i$ and $\boldsymbol{t}_i$ both contain $1024$ neurons.}
\label{tab:mnist-n-layers-asso-loss}
\centering
\begin{tabular}{l||lll}
\hline
Number of component layers                           & 1 layer                 & 2 layers                & 3 layers  \\
\hline\hline
$||\boldsymbol{s}_1'-\boldsymbol{t}_1||_2^2$ & $1.2488 \times 10^{-5}$ & $1.5469 \times 10^{-5}$ & $1.2219 \times 10^{-5}$ \\
$||\boldsymbol{s}_2'-\boldsymbol{t}_2||_2^2$ & -                       & $3.5818 \times 10^{-7}$ & $3.8033 \times 10^{-7}$ \\
$||\boldsymbol{s}_3'-\boldsymbol{t}_3||_2^2$ & -                       & -                       & $6.7192 \times 10^{-10}$\\
\hline
\end{tabular}
\end{table}

\begin{table}
\caption{Number of layers vs. the training accuracy and vs. the test accuracy on the MNIST
dataset after 200 epochs. Referring to Figure~\ref{fig:asso-learn-all}, for each layer,
the corresponding $\boldsymbol{s}_i$ and $\boldsymbol{t}_i$ both contain $1024$ neurons.
The bridge layer in the top layer includes $5120$ neurons.}
\label{tab:mnist-n-layers-acc-cmp}
\centering
\begin{tabular}{l||ccc}
\hline
Number of component layers  & 1 layer & 2 layers & 3 layers \\
\hline\hline
Training accuracy & 1.0     & 1.0      & 1.0      \\
Test accuracy     & 0.9849  & 0.9860   & 0.9871   \\
\hline
\end{tabular}
\end{table}

This section presents the results of experiments with different numbers of
component layers on the MNIST dataset. For each component layer $i$, both the
corresponding $\boldsymbol{s}_i$ and $\boldsymbol{t}_i$ have $1024$ neurons,
and $\boldsymbol{s}'_{\ell}$ (i.e., the output of the bridge at the top
layer) contains $5120$ neurons.

First, we show that each component indeed incrementally refines the associated
loss of the one immediately below it. Specifically, we applied AL to the MLP and
experimented with different numbers of component layers. As shown in
Table~\ref{tab:mnist-n-layers-asso-loss}, adding more layers truly decreases
the associated loss, and the associated loss at an upper layer is smaller than
that at a lower layer.

Second, we show that adding more layers helps transform
$\boldsymbol{x}$ into $y$. As shown in Table~\ref{tab:mnist-n-layers-acc-cmp},
adding more layers increases the test accuracy.

\subsection{Metafeature Visualization and Quantification}

\begin{figure}[ht]
\begin{center}
\centerline{\includegraphics[width=5in]{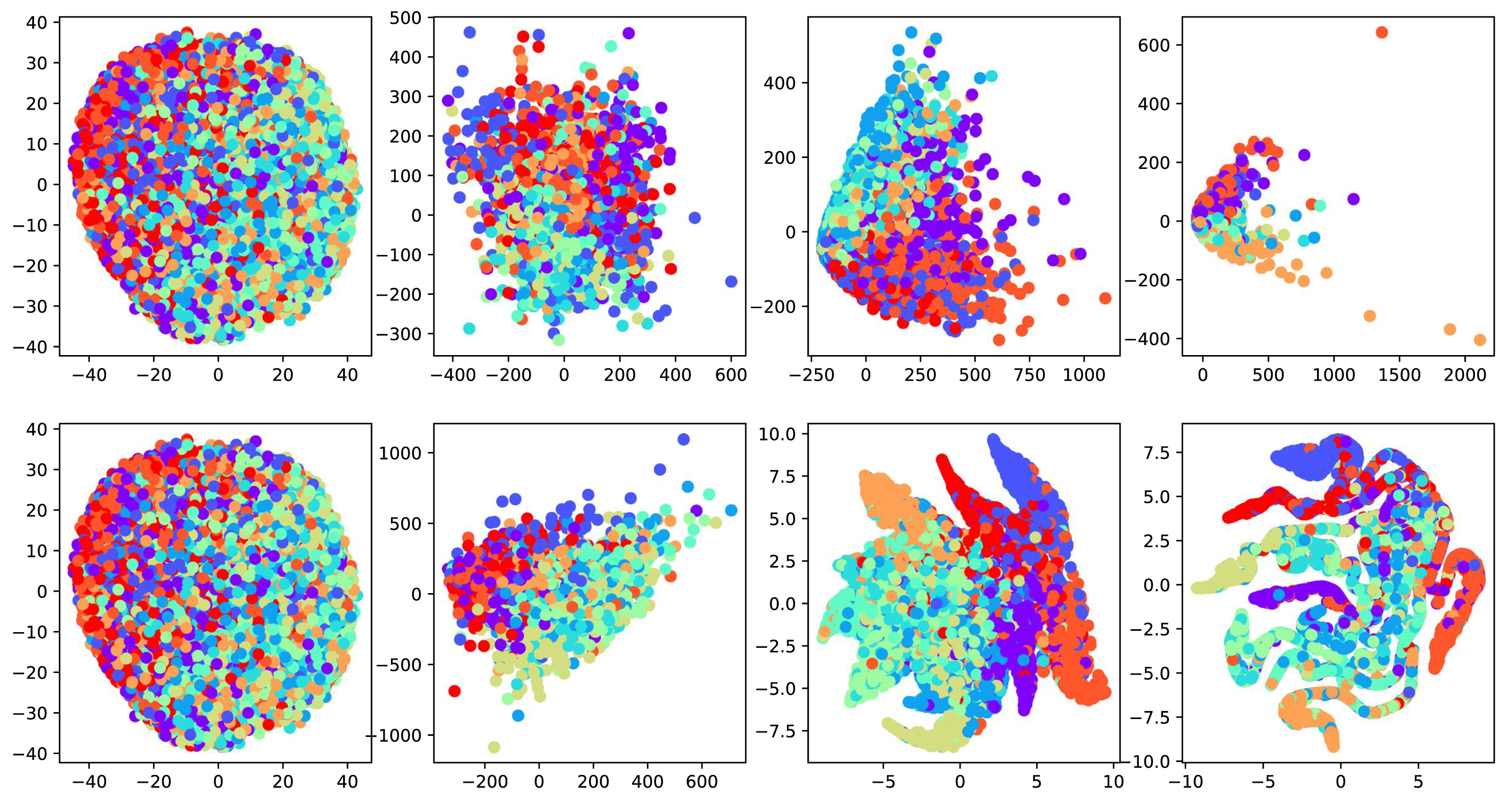}}
\caption{t-SNE visualization of the MLP on the CIFAR-10 dataset. The different colors represent different labels.
The figures in the first row are the results of the raw data, $2^{\textrm{nd}}$ layer,  $4^{\textrm{th}}$ layer, and output layer when using BP.
The second row shows the corresponding results for AL.}
\label{fig:CIFAR10tsneMLP}
\end{center}
\end{figure}

\begin{figure}[ht]
\begin{center}
\centerline{\includegraphics[width=5in]{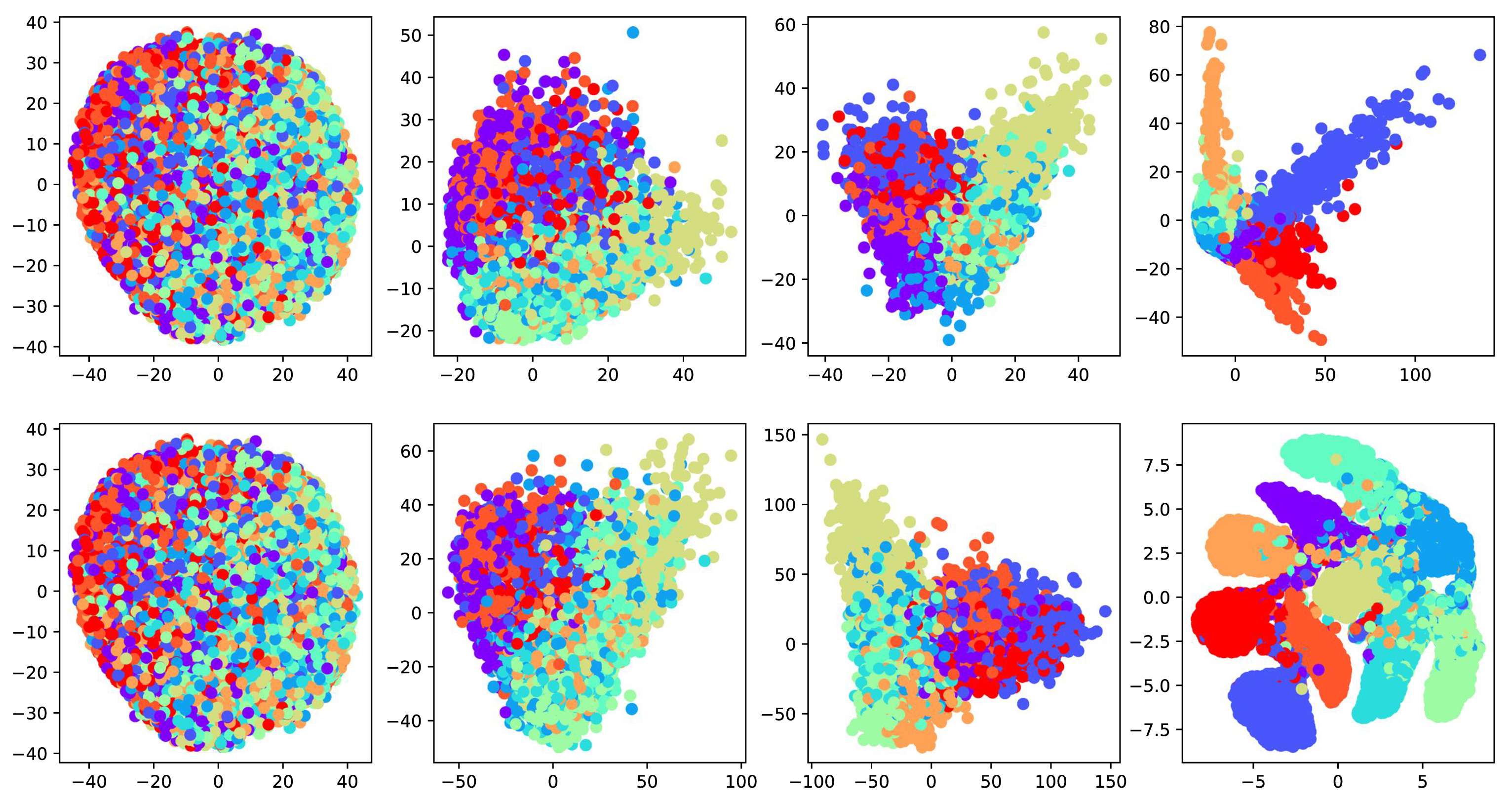}}
\caption{t-SNE visualization of Vanilla CNN with CIFAR-10 dataset. The different colors represent different labels. The figures in the first row are the results 
of the raw data, $4^{\textrm{th}}$ layer, $8^{\textrm{th}}$ layer, and $12^{\textrm{th}}$ layers when using BP. The second row shows the corresponding results for AL.}
\label{fig:CIFAR10tsneCNN}
\end{center}
\end{figure}

\begin{table}
\caption{A comparison of the inter- and intraclass distances and the ratio of
the two. We highlight the winner in bold font.}
\label{tab:inter-intra-dist}
\centering
\resizebox{\textwidth}{!}{
\begin{tabular}{l||l|l|rrr}
\hline
Dataset                    & Network                      & Method & \multicolumn{1}{l}{Interclass distance} & \multicolumn{1}{l}{Intraclass distance} & \multicolumn{1}{l}{Inter:Intra ratio} \\
\hline\hline
\multirow{4}{*}{CIFAR-10}  & \multirow{2}{*}{MLP} & BP & $39.36$ & $67.97$ & $0.58$ \\
                           & & AL & $0.73$ & $0.66$ & $\boldsymbol{1.11}$ \\
\cline{2-6} 
                           & \multirow{2}{*}{Vanilla CNN} & BP & $41.82$ & $26.87$ & $1.56$ \\
                           & & AL & $1.17$ & $0.36$ & $\boldsymbol{3.25}$ \\
\hline
\multirow{4}{*}{CIFAR-100} & \multirow{2}{*}{MLP} & BP & $114.42$ & $342.65$ & $0.33$ \\
                           & & AL & $0.23$ & $0.28$ & $\boldsymbol{0.82}$ \\
\cline{2-6} 
                           & \multirow{2}{*}{Vanilla CNN} & BP & $114.71$ & $163.43$ & $0.70$ \\
                           & & AL & $0.55$ & $0.51$ & $\boldsymbol{1.08}$ \\
\hline
\end{tabular}
}
\end{table}

To determine whether the hidden layers truly learn useful metafeatures when
using AL, we used t-SNE~\citep{maaten2008visualizing} to visualize
the $2^{\textrm{nd}}$, $4^{\textrm{th}}$ hidden layers and output layer in the 6-layer MLP model
and the $4^{\textrm{th}}$, $8^{\textrm{th}}$, and $12^{\textrm{th}}$hidden layers in the 14-layer Vanilla
CNN model on the CIFAR-10 dataset. For comparison purposes, we also visualize
the corresponding hidden layers trained using BP. As shown in
Figure~\ref{fig:CIFAR10tsneMLP} and Figure~\ref{fig:CIFAR10tsneCNN}, the
initial layers seem to extract less useful metafeatures than the later layers because the labels are
difficult to distinguish in the corresponding figures. However, a comparison of
the last few layers shows that AL groups the data points of the
same label more accurately than BP, which suggests that AL likely learns
better metafeatures.

To assess the quality of the learned metafeatures, we calculated the intra- and
interclass distances of the data points based on the metafeatures. We
computed the intraclass distance $d_k^{intra}$ as the average distance between
any two data points in class $k$ for each class. The interclass distance is
the average distance between the centroids of the classes. We also computed the
ratio between inter- and intraclass distance to determine the quality of the
metafeatures generated by AL and
BP~\citep{michael1973experimental, luo2019g}. As shown in
Table~\ref{tab:inter-intra-dist}, AL performs better than
BP on both the CIFAR-10 and CIFAR-100 datasets because AL
generates metafeatures with a larger ratio between inter- and intraclass
distance.

\section{Discussion and Future Work} \label{sec:disc}

Although BP is the cornerstone of today's deep learning algorithms, it is far
from ideal, and therefore, improving BP or searching for 
alternatives is an important research direction. This paper discusses
AL, a novel process for training deep neural networks without
end-to-end BP. Rather than calculating gradients in a layerwise
fashion based on BP, AL removes the dependencies
between the parameters of different subnetworks, thus allowing each subnetwork
to be trained simultaneously and independently. Consequently, we may utilize
pipelines to increase the training throughput. Our method is biologically
plausible because the targets are local and the gradients are not obtained
from the output layer. Although AL does not directly minimize the
prediction error, its test accuracy is comparable to, and sometimes better than,
that of BP, which does directly attempt to minimize the prediction
error. Although recent studies have begun to use local losses instead of
backpropagating the global loss~\citep{nokland2019training}, these local losses
are computed mainly based on (or are at least partially based on) the difference
between the target variable and the predicted results. Our method is unique because
in AL, most of the layers do not interact with the target variable.

Current strategies to parallelize the training of a deep learning model usually
distribute the training data into different computing units and aggregate
(e.g., by averaging) the gradients computed by each computing unit. Our work, on
the other hand, parallelizes the training step by computing the parameters of the
different layers simultaneously. Therefore, AL is not an
alternative to most of the other parallel training approaches but can integrate
with the abovementioned approach to further improve the training throughput.

Years of research have allowed us to gradually understand the proper hyperparameter
settings (e.g., network structure, weight initialization, and activation function) when training a neural network based on BP. However, these
settings may not be appropriate when training by AL. Therefore,
one possible research direction is to search for the right settings for this
new approach.

We implemented AL in TensorFlow. However, we were unable to
implement the ``pipelined'' AL that was shown in Table~\ref{tab:pipeline}
within a reasonable period because of the technical challenges of task
scheduling and parallelization in TensorFlow. We decided to leave this part as
future work. However, we ensure that the gradients propagate only within each
component, 
so theoretically, a pipelined AL should be able to be implemented.

Another possible future work is validating AL on other datasets.
(e.g., ImageNet, Microsoft Common Objects in Context (MS COCO), and Google's Open Images) and even on datasets unrelated to computer vision, such as those used in signal processing, natural language processing, and recommender systems. Yet another future work
is the theoretical work of AL, as this may help us understand why
AL outperforms BP under certain network structures.
In the longer term, we are highly interested in investigating optimization
algorithms beyond BP and gradients.

\subsection*{Acknowledgments}

We acknowledge partial support by the Ministry of Science and
Technology under grant no. MOST 107-2221-E-008-077-MY3.
We thank the reviewers for their informative feedback.

\bibliographystyle{apa}
\bibliography{ref}
\end{document}